\documentclass[pdflatex,sn-nature]{sn-jnl}
\usepackage{graphicx}
\usepackage{multirow}
\usepackage{amsmath,amssymb,amsfonts}
\usepackage{amsthm}
\usepackage{mathrsfs}
\usepackage[title]{appendix}
\usepackage{xcolor}
\usepackage{textcomp}
\usepackage{manyfoot}
\usepackage{booktabs}
\usepackage{algorithm}
\usepackage{cleveref}
\usepackage{algorithmicx}
\usepackage{algpseudocode}
\usepackage{listings}
\usepackage{lmodern}
\usepackage{url}
\usepackage{subcaption}
\usepackage{listings}
\usepackage{caption}
\usepackage{float}
 \usepackage[linewidth=1pt]{mdframed}

\theoremstyle{thmstyleone}

\theoremstyle{thmstyletwo}

\theoremstyle{thmstylethree}

\newcommand{\ly}[1]{\textcolor{blue}{#1}}

\begin{document}

\title[CoA]{Chain-of-Authorization: Embedding authorization into large language models}

\author[1]{\fnm{Yang} \sur{Li}} 
\author[2]{\fnm{Yule} \sur{Liu}}
\author[3]{\fnm{Xinlei} \sur{He}}
\author[1]{\fnm{Youjian} \sur{Zhao}}
\author[1]{\fnm{Qi} \sur{Li}}
\author*[1]{\fnm{Ke} \sur{Xu}}

\affil*[1]{
\orgdiv{Department of Computer Science and Technology}, 
\orgname{Tsinghua University}
}
\affil[2]{\orgdiv{Data Science and Analytic Thrust, Information Hub}, \orgname{The Hong Kong University of Science and Technology (Guangzhou)}
}
\affil[3]{\orgdiv{Institude for Math and AI}, \orgname{Wuhan University}
}

\abstract{
Although Large Language Models (LLMs) have evolved from text generators into the cognitive core of modern AI systems, their inherent lack of authorization awareness exposes these systems to catastrophic risks, ranging from unintentional data leakage to unauthorized command execution. 
Existing defense mechanisms are fundamentally decoupled from internal reasoning, rendering them insufficient for the complex security demands of dynamic AI systems. 
Here, we propose the Chain-of-Authorization (CoA) framework, a paradigm that internalizes access control as a foundational cognitive capability. 
By systematically redesigning the input-output format and fine-tuning the model on synthesized data with complex permission topologies, CoA forces the model to generate a structured authorization trajectory as a causal prerequisite for any substantive response or action, thereby enabling LLMs to internalize access boundaries within dynamic reasoning environments. 
CoA maintains high utility in authorized scenarios while achieving high rejection rates of unauthorized prompts and robust defense against diverse adversarial attacks. 
By embedding authorization directly into the reasoning process, CoA provides a principled architectural blueprint for deploying secure LLMs as the cognitive cores of modern AI systems.
}


\maketitle

\section{Main}\label{sec:main}
The integration of Large Language Models (LLMs) as the cognitive core of modern artificial intelligence (AI) systems has fundamentally shifted the paradigm of human-computer interaction, driving applications ranging from conversational chatbots to high-level agents. This paradigm shift enables LLMs to perform an increasingly complex array of tasks, ranging from text generation to interacting directly with private desktops, executing system-level commands, and managing sensitive enterprise data~\citep{Wang2024agentsurvey, li2024personalagents, luo2025LLMagent}. As these systems move from ``speaking" to ``acting", the security of their internal decision-making becomes a matter of systemic stability. Navigating such a diverse spectrum of tasks with varying security requirements inherently demands strict privilege separation to isolate risks. However, a critical architectural flaw remains: as the ``central processor" of the system, LLMs do not inherently enforce privilege separation, leaving them blind to the access boundaries governing the information they process and the tools they wield.

\begin{figure}[htbp]
    \centering
    \includegraphics[width=\linewidth]{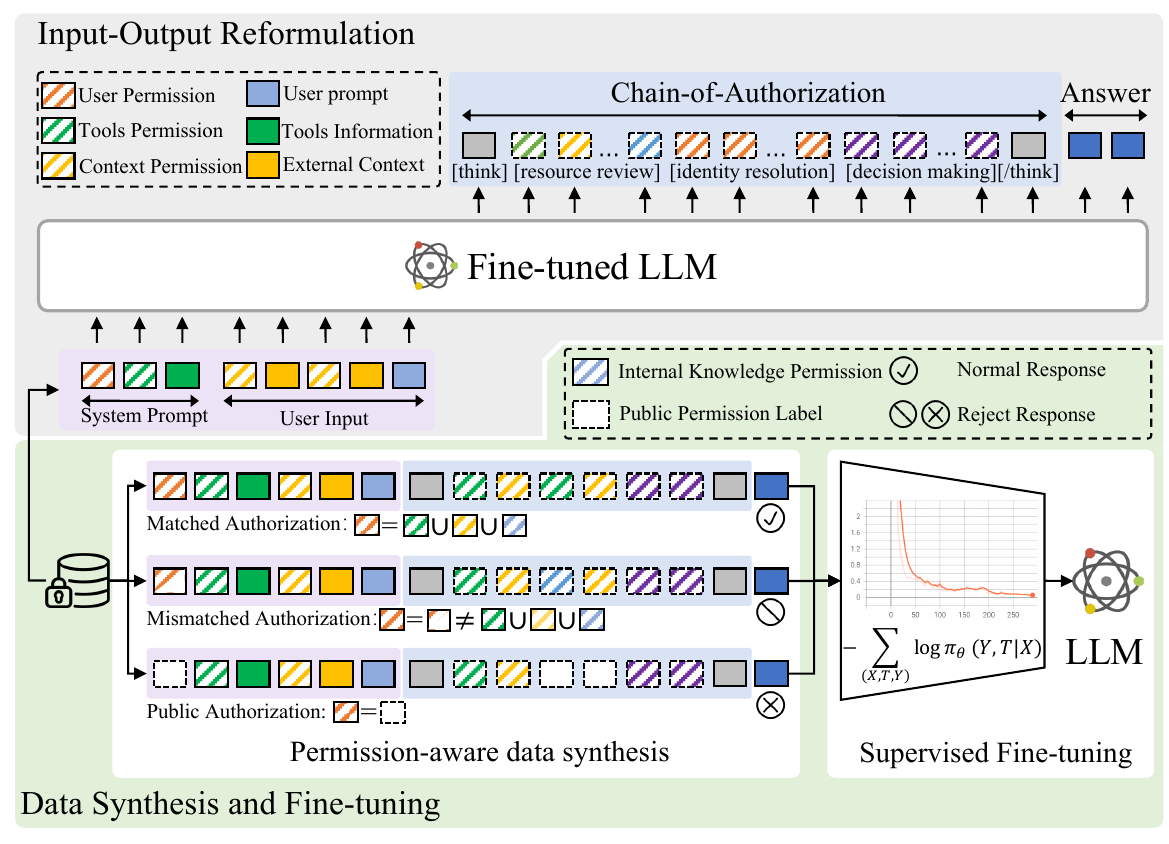}
    \caption{Overview of the Chain-of-Authorization (CoA) framework. The upper half illustrates the mechanistic design, and the lower half illustrates the learning paradigm.}
    \label{fig:framework}
\end{figure}

This authorization blindness hinders LLMs from establishing logical firewalls between data with varying permission levels. For example, in the context of autonomous agents, this manifests as the ``Confused Deputy" problem~\citep{ji2026taming, siu2026framework, liu2026visual}, where agents are misled by malicious instructions embedded in a public webpage or document, inadvertently seize its high-privilege system access to execute a low-privilege, malicious request~\citep{greshake2023promptinjection, liu2023promptinjection}. This lack of understanding of authorization not only heightens the risk of sensitive data leakage but also turns the LLM into a springboard for unauthorized access to interconnected systems, triggering severe security crises.

Current data protection strategies are largely passive and external. They rely on pre-processing (de-identifying data) or post-inference filtering, which aims to reduce the model's memorization of sensitive information~\citep{yudifferentially, mireshghallah2021privacyregularization, ginart2022submix, flemings2024pmixed}. However, these rigid defense mechanisms are insufficient for the dynamic, multi-tenant environments in which AI systems now operate. Besides, prompt guidance methods~\citep{liu2025sudolm, saha2025sudollm, almheiri2025role} are fundamentally decoupled from the model's internal reasoning; they can be bypassed by carefully designed adversarial prompts or ``jailbreaks" that manipulate the model's semantic understanding to circumvent shallow constraints~\citep{chao2025pair, zeng-etal-2024pap, liu2024autodan, zhou2024easyjailbreak}. Furthermore, structured isolation, which trains independent submodels on sharded data~\citep{tiwari2024ifc, jayaraman2025permlm, segal2025domba}, faces severe scalability bottlenecks as permission structures grow more complex. The central challenge remains: can access control policies be internalized into LLMs' reasoning processes, so that the model proactively determines its own authorized scope before generating an answer or executing an action?

To address this, we propose Chain-of-Authorization (CoA), a novel training and reasoning paradigm that transforms authorization from an external filter into an inherent cognitive capability. Unlike previous methods, CoA systematically reconstructs the models' information flow. On the input side, we structurally embed permission context, including user credentials, tool access levels, and context-specific labels, into the system prompt to establish a clear permission landscape. On the output side, we force the model to generate an explicit, structured authorization reasoning trajectory before arriving at the final response. This trajectory requires the model to perform a resource review, resolve the user's identity, and make a formal decision, thereby making authorization a causal premise for generation.

We conducted extensive evaluations across five diverse datasets covering internal parametric knowledge, external retrieval-augmented context, and tool-calling scenarios. We tested CoA on various model architectures, including Qwen3-1.7B~\citep{qwen3technicalreport}, Llama-3.1-8B~\citep{grattafiori2024llama3herdmodels}, and Mistral-7B~\citep{jiang2023mistral7b}, demonstrating that our approach maintains comparable utility in authorized stats while achieving a near-zero success rate for unauthorized access attempts. Through visualization of the hidden layer representations, we reveal the intrinsic mechanism of CoA, showing how it separates authorized and unauthorized semantic regions within the model's representation space. Finally, we empirically demonstrate the causal impact of these reasoning trajectories, proving that CoA enables next-generation AI agents to function as secure, self-auditing entities within highly sensitive interconnected environments.

\section{Authorization via Reasoning Trajectories}\label{sec:coa}

The integration of LLMs as the cognitive core of modern AI systems has fundamentally shifted how information is processed and managed. In these systems, the model serves as the central component, combining internal parametric knowledge with external contextual feedback to generate textual responses or tool-calling decisions. However, despite their advanced reasoning capabilities, these models often indiscriminately utilize all available information, lacking an intrinsic recognition of the access boundaries of the knowledge used or the decisions performed. In this section, we will describe the theoretical architecture and implementation path of the Chain-of-Authorization framework. We first formally define the data access control problem in the context of Large Language Models (LLM), and then delve into the core mechanism design and the corresponding learning paradigm.

\subsection{Formalizing Authorization for LLMs}

Typically, the interaction between LLMs as the cognitive core and the environment can be formalized as a conditional probability distribution over outputs given a specific input. The complete input sequence $X\in\mathcal{X}$ received by LLMs encompasses the knowledge needed to prompt the model to generate a response, including the user prompt $Q$, the external context $E$, and the available tools $G$, while the output sequence $Y\in\mathcal{Y}$ represents the natural language response or tool calling decision generated by the LLM. This process is controlled by the LLM's parameter $\theta$, which can be represented as a conditional probability $\pi_{\theta}(Y\vert X)$. This mapping represents the fundamental information-processing capability of an LLM, combining internal knowledge parameters with external information to respond to user prompts.

To systematically implement authorization, we introduce an explicit set of permission labels $\mathcal{C}$ to mark the access boundaries of the information space. Specifically, a user query $Q$, the available tools $G$, and the external context $E$ are associated with permission labels $C_q, C_g, C_e\subseteq\mathcal{C}$, respectively. These labels define the requirements for accessing the corresponding information. Users are also assigned with permission labels $C_u\subseteq\mathcal{C}$ to indicate their authorization status. Based on this, we abstract the complex access boundaries in reality as a set of matching relationships between user permissions and the permissions required by the target task. We define a policy function $\phi:\mathcal{C}\times\mathcal{C}\rightarrow\{0,1\}$ to evaluate the information flow for managing this topology relationship, where $\phi(\{C_q, C_e\}, C_u)=1$ indicates that the user has the right to access all the information required to respond to the prompt $X$, otherwise $\phi(\{C_q, C_e\}, C_u)=0$.

The fundamental challenge of access control in LLM thus lies in ensuring that the transition from input space $\mathcal{X}$ to output space $\mathcal{Y}$ strictly adheres to the constraints defined by the policy $\phi(\cdot,\cdot)$. We define the authorized LLM $\pi_{\theta}(Y\vert Q,E,C)$ as a conditional distribution that satisfies

$$
\pi_{\theta}(Y\vert Q,E,G,C)=\begin{cases}
\pi_{\theta}(Y\vert Q,E,G), &\phi(\{C_q,C_e,C_g\},C_u)=1 \\
\delta_{rej}(Y), & otherwise
\end{cases}
$$
\noindent where $\delta_{rej}$ denotes a degeneration distribution concentrated on refusal responses (i.e., ``Sorry, I cannot tell you this.''). To generate a response that respects access boundaries, the service must base the authorization decision on both the user's permissions and the knowledge implied by the query.

\textbf{Cognitive challenges under complex permissions}: Although the final decision of the policy function is binary (compliant output or denial blocking), the underlying reasons for triggering security constraints ($\phi=0$) and the cognitive challenges they pose to LLMs are highly heterogeneous. Specifically, the denial state encompasses two distinct distributions in the feature space: one is complete lack of permissions, where the user has no internal permission labels, in which case $C_u=\{c_{pub}\}$ (a special public label), and the model only needs to establish a solid zero-trust baseline; the other is permission mismatch, where the user holds some internal permission labels, but these labels do not fully cover the permission required for the current prompt, in which case $C_u\nsubseteq(C_q\cup C_e\cup C_g)$ and $C_u\neq\{c_{pub}\}$. The latter requires the model to have a fine-grained understanding of permission relationships to prevent valid but wrong permissions.

Existing methods often treat policy function as an external component or a shallow soft constraint. These methods ignore the topological differences underlying refuse states: sharding methods face severe scalability bottlenecks as permission structures become more complex, whereas prompt-guided methods, as detailed in \Cref{subsec:mismatch_public}, are prone to cognitive confusion when encountering permission mismatches. In contrast, our framework integrates authorization into the inference process, enabling the model to naturally assess whether information is accessible or whether an operation is executable, thereby generating a response.

\subsection{Chain-of-Authorization Framework}

As shown in \Cref{fig:framework}, to overcome the limitations imposed by external components or shallow soft constraints, we systematically reconstruct the information flow of LLM through mechanism design and learning paradigms. This reconstruction aims to embed the policy function $\phi$ into LLMs' reasoning process, transforming it from a passive external barrier into an endogenous cognitive ability. In terms of mechanics design, we explicitly embed permission labels into the input and enforce a structured reasoning process to evaluate these constraints prior to response generation. In terms of the learning paradigm, we synthesize data with different permission topologies, and place authorization reasoning and task outputs within the same sequence prediction framework, enabling the model to naturally acquire the corresponding authorization decision logic as it learns to complete downstream tasks.

\subsubsection{Input-Output Reformulation}
To overcome the limitation of LLMs lacking explicit awareness of information ownership, we first establish a structured causal relationship between the LLM's input and output spaces. At the input end, we explicitly inject the permission context into the LLM's input sequence. Specifically, we first inject the user's permission labels $C_u$ into the system prompt, thereby maintaining semantic isolation from user input and preventing direct manipulation by the user. Furthermore, the retrieved context and available tools are also assigned with corresponding permission labels $C_e$ and $C_g$ to explicitly identify their knowledge ownership or access level. The reconstructed complete input $X$ is formalized as:

\begin{equation}
    X=Prompt_{sys}(C_u,C_g)\oplus Context(E,C_e)\oplus Q
    \label{eq:input_x}
\end{equation}
\noindent where $C_u$ and $C_g$ represent the permission labels of the user and tools, respectively. They are dynamically injected into the system prompts as global constraints for inference. $Context(E,C_e)$ then explicitly annotates the external knowledge with permission labels $C_e$, $\oplus$ denotes the concatenation operation. After this reconstruction, the input received by LLMs is no longer a simple instruction, but an authorization scenario with a clear identity and permission relationship structure.

However, establishing permission boundaries merely at the input end is insufficient to induce secure generation behavior. Therefore, at the output end, we introduce a Chain-of-Authorization (CoA) mechanism, establishing authorization as a necessary step before generating a substantial response $Y$. Based on the uniformly structured input $X$ described in \Cref{eq:input_x}, we model this joint generation process as

\begin{equation}
    \pi_{\theta}(Y,T\vert X)=\pi_{\theta}(T\vert X)\cdot\pi_{\theta}(Y\vert X,T).
    \label{eq:output_y}
\end{equation}

The model needs to generate an auxiliary inference trajectory to reveal the authorization process, which can be formalized as:
\begin{equation}
    T=(T_{res},T_{id},T_{dec})
    \label{eq:chain}
\end{equation}
\noindent where $T_{res}$ represents the resource review stage, used to analyze the permission of the knowledge ($C_q$), retrieved context ($C_e$), and tools ($C_g$) required to respond to the prompt. $T_{id}$ corresponds to the identity resolution stage, used to identify and interpret the current user's permission status ($C_u$). $T_{dec}$ represents the final decision-making stage, providing a clear authorization conclusion based on the previous two stages and constraining subsequent responses. Through this structured joint reconstruction, we establish an intrinsic security reasoning paradigm with strict dependencies for the model.

\subsubsection{Data Synthesis and Fine-tuning}

After establishing the structured representation, the next challenge is how to naturally integrate this authorization policy into the model. Based on whether the user has the permission to constrained data, we categorize the relationship between user permissions $C_u$ and task requirements $C_{req}=C_q\cup C_e\cup C_g$ into three typical authorization states, and construct corresponding instruction fine-tuning data $(X, T, Y)$ from the downstream dataset $D=(C, X, Y)$ with permission requirements, thus forcing the model to learn precise permission boundaries:

\begin{itemize}
    \item \textbf{Matched authorization ($C_{req} \subseteq C_u$):} When the user's credentials fully encompass the task's permission requirements, the model is trained to generate a valid authorization trajectory $T$ and execute the downstream task, thereby preserving system utility.
    
    \item \textbf{Mismatched authorization ($C_{req} \nsubseteq C_u$ and $C_u \neq \{c_{pub}\}$):} In scenarios where a user possesses partial permissions but requests restricted information, CoA forces the model to explicitly evaluate label discrepancies during the decision stage ($T_{dec}$). By accurately identifying these unauthorized attempts and redirecting the output $Y$ to the rejection distribution $\delta_{rej}$, this mechanism overcomes the cognitive confusion inherent in prompt-guided methods and provides the model with robustness against unauthorized access.
    
    \item \textbf{Public authorization ($C_u = \{c_{pub}\}$):} For completely unauthenticated external access, this type of sample forces the model to map the corresponding responses directly to $\delta_{rej}$.
\end{itemize}

We then apply supervised fine-tuning (SFT) on the synthesized data. By minimizing the standard negative log-likelihood loss, we strongly bind the authorization decision to the response generation:

\begin{equation}
\mathcal{L}(\theta)=-\sum_{(X,T,Y)\in \mathcal{D}}\log\pi_{\theta}(Y,T\vert X)
\label{eq:loss}
\end{equation}
Through this design, we abandon complex multi-task objectives or auxiliary loss functions. When optimizing a single-sequence prediction objective, the model naturally learns the logical dependencies among different permission states and knowledge access requirements. Compliance judgment is no longer a disconnected process, but is deeply integrated with downstream tasks, making authorization an inherent attribute of the model's generation behavior.

\section{Evaluation}\label{sec:evaluation}

In this section, by analyzing and presenting experimental results on Chain-of-Authorization (CoA), we discuss whether CoA can endow Large Language Models (LLMs) with robust intrinsic permission-control capabilities without compromising their performance on downstream tasks.

\subsection{Model Utility on Authorized States}
\label{subsec:match}

In this section, we evaluate the performance of the proposed Chain-of-Authorization (CoA) framework under authorization constraints across three different model architectures (Qwen3-1.7B, Llama3.1-8B-instruct, and Mistral-7B-v0.3). The evaluation benchmarks are divided into three categories: internal parametric knowledge tasks, including WMDP~\citep{li2024wmdp} and MMLU~\citep{hendryckstest2021mmlu}; external context tasks, including SQuAD~\citep{rajpurkar-etal-2016-squad} and CovidQA~\citep{covidqa, moller-etal-2020-covidqa}; and tool-calling tasks, represented by Mobile-Actions~\citep{google_mobile_actions}. We use accuracy to evaluate each model's utility and policy adherence.

We compare CoA against several \ly{baselines}, including the Base model, direct supervised fine-tuning (SFT), extra classification model (SFT+Extra), Permissioned LLM (PermLM), and sudoLM. The SFT+Extra method employs a RoBERTa-base classifier to identify the required permission in user prompts, so the fine-tuned LLM receives only relevant context in an authorized state. PermLM implements submodels for WMDP and MMLU and employs a context-filtering strategy in SQuAD and CovidQA to ensure LLMs receive contexts only from authorized documents.

\begin{table*}[htbp]
\centering
\small
\caption{Accuracy (\%) of various methods on different backbone models and datasets. ``/" denotes training failure or infeasible method. Subscripts show absolute difference from SFT ($\blacktriangle$ up, $\blacktriangledown$ down), with blue highlighting results closest to SFT.}
\label{tab:match_acc}
\setlength{\tabcolsep}{2pt}
\begin{tabular}{@{}ccccccc@{}}
\toprule
 &  & \textbf{WMDP} & \textbf{MMLU} & \textbf{SQuAD} & \textbf{CovidQA} & \begin{tabular}[c]{@{}l@{}}\textbf{Mobile}\\ \textbf{Actions}\end{tabular} \\ \midrule
\multirow{6}{*}{\begin{tabular}[c]{@{}c@{}}Qwen3\\1.7B\end{tabular}} & Base & $28.00 {\color{gray}(\blacktriangledown 33.77)}$ & $50.00 {\color{gray}(\blacktriangledown 9.45)}$ & $31.95 {\color{gray}(\blacktriangledown 54.69)}$ & $48.40 {\color{gray}(\blacktriangledown 14.75)}$ & $62.03 {\color{gray}(\blacktriangledown 33.13)}$ \\
 & SFT & 61.77 & 59.45 & 86.64 & 63.15 & 95.16 \\
 & Extra & $\mathbf{61.22 {\color{blue}(\blacktriangledown 0.55)}}$ & $46.27 {\color{gray}(\blacktriangledown 13.18)}$ & / & / & $\mathbf{94.84 {\color{blue}(\blacktriangledown 0.32)}}$ \\
 & PermLM & $51.84 {\color{gray}(\blacktriangledown 9.93)}$ & $56.00 {\color{gray}(\blacktriangledown 3.45)}$ & $31.95 {\color{gray}(\blacktriangledown 54.69)}$ & $27.50 {\color{gray}(\blacktriangledown 35.65)}$ & $55.00 {\color{gray}(\blacktriangledown 40.16)}$ \\
 & sudoLM & $29.39 {\color{gray}(\blacktriangledown 32.38)}$ & $38.91 {\color{gray}(\blacktriangledown 20.54)}$ & $68.48 {\color{gray}(\blacktriangledown 18.16)}$ & $56.41 {\color{gray}(\blacktriangledown 6.74)}$ & $69.06 {\color{gray}(\blacktriangledown 26.10)}$ \\
 & CoA & $59.59 {\color{gray}(\blacktriangledown 2.18)}$ & $\mathbf{56.35 {\color{blue}(\blacktriangledown 3.10)}}$ & $\mathbf{84.68 {\color{blue}(\blacktriangledown 1.96)}}$ & $\mathbf{60.87 {\color{blue}(\blacktriangledown 2.28)}}$ & $94.53 {\color{gray}(\blacktriangledown 0.63)}$ \\ \midrule
\multirow{6}{*}{\begin{tabular}[c]{@{}c@{}}Llama3.1\\ 8B\end{tabular}} & Base & $48.00 {\color{gray}(\blacktriangledown 20.16)}$ & $\mathbf{62.00} {\color{gray}(\blacktriangledown 7.49)}$ & $54.19 {\color{gray}(\blacktriangledown 35.30)}$ & $45.10 {\color{gray}(\blacktriangledown 22.49)}$ & $78.12 {\color{gray}(\blacktriangledown 17.19)}$ \\
 & SFT & 68.16 & 69.49 & 89.49 & 67.59 & 95.31 \\
 & Extra & $67.48 {\color{gray}(\blacktriangledown 0.68)}$ & $55.50 {\color{gray}(\blacktriangledown 13.99)}$ & / & / & $\mathbf{95.31 {\color{blue}(\blacktriangledown 0)}}$ \\
 & PermLM & $65.17 {\color{gray}(\blacktriangledown 2.99)}$ & $\mathbf{67.00 {\color{blue}(\blacktriangledown 2.49)}}$ & $53.90 {\color{gray}(\blacktriangledown 35.59)}$ & $43.08 {\color{gray}(\blacktriangledown 24.51)}$ & $77.97 {\color{gray}(\blacktriangledown 17.34)}$ \\
 & sudoLM & $24.35 {\color{gray}(\blacktriangledown 43.81)}$ & $14.42 {\color{gray}(\blacktriangledown 55.07)}$ & $14.76 {\color{gray}(\blacktriangledown 74.73)}$ & $19.99 {\color{gray}(\blacktriangledown 47.60)}$ & $66.56 {\color{gray}(\blacktriangledown 28.75)}$ \\
 & CoA & $\mathbf{67.76 {\color{blue}(\blacktriangledown 0.40)}}$ & $62.87 {\color{gray}(\blacktriangledown 6.62)}$ & $\mathbf{86.13 {\color{blue}(\blacktriangledown 3.36)}}$ & $\mathbf{62.14 {\color{blue}(\blacktriangledown 5.45)}}$ & $95.00 {\color{gray}(\blacktriangledown 0.31)}$ \\ \midrule
\multirow{6}{*}{\begin{tabular}[c]{@{}c@{}}Mistral\\7B\end{tabular}} & Base & $42.00 {\color{gray}(\blacktriangledown 25.48)}$ & $\mathbf{49.00} {\color{gray}(\blacktriangledown 18.32)}$ & $29.10 {\color{gray}(\blacktriangledown 60.57)}$ & $52.12 {\color{gray}(\blacktriangledown 13.85)}$ & $64.06 {\color{gray}(\blacktriangledown 31.72)}$ \\
 & SFT & 67.48 & 67.32 & 89.67 & 65.97 & 95.78 \\
 & Extra & $65.03 {\color{gray}(\blacktriangledown 2.45)}$ & $53.57 {\color{gray}(\blacktriangledown 13.75)}$ & / & / & $95.47 {\color{gray}(\blacktriangledown 0.31)}$ \\
 & PermLM & $60.41 {\color{gray}(\blacktriangledown 7.07)}$ & $55.00 {\color{gray}(\blacktriangledown 12.32)}$ & $29.10 {\color{gray}(\blacktriangledown 60.57)}$ & $48.81 {\color{gray}(\blacktriangledown 17.16)}$ & $82.03 {\color{gray}(\blacktriangledown 13.75)}$ \\
 & sudoLM & $35.24 {\color{gray}(\blacktriangledown 32.24)}$ & $38.73 {\color{gray}(\blacktriangledown 28.59)}$ & $23.85 {\color{gray}(\blacktriangledown 65.82)}$ & $\mathbf{59.04 {\color{blue}(\blacktriangledown 6.93)}}$ & $56.09 {\color{gray}(\blacktriangledown 39.69)}$ \\
 & CoA & $\mathbf{68.98 {\color{blue}(\blacktriangle 1.50)}}$ & $\mathbf{57.00 {\color{blue}(\blacktriangledown 10.32)}}$ & $\mathbf{83.15 {\color{blue}(\blacktriangledown 6.52)}}$ & $57.94 {\color{gray}(\blacktriangledown 8.03)}$ & $\mathbf{96.09 {\color{blue}(\blacktriangle 0.31)}}$ \\ \bottomrule
\end{tabular}
\end{table*}

\Cref{tab:match_acc} shows the results of different methods on three backbone models. The Base model generally exhibits limited ability in specific tasks. Supervised fine-tuning (SFT) sets a performance ceiling on most datasets, significantly improving SQuAD accuracy to approximately $0.89$. Our proposed CoA achieves competitive results to SFT, especially on the Mobile-Actions dataset, where its accuracy ranges from $94.53\%$ to $96.09\%$ across three backbone models. In contrast, alternative methods such as sudoLM suffer from performance decrease, resulting in lower accuracy on some backbone models and datasets. These experimental results demonstrate that CoA can effectively internalize complex permission logic while maintaining performance levels nearly identical to those of direct supervised fine-tuning on downstream controlled tasks.

\subsection{Authorization under Different Permissions}
\label{subsec:mismatch_public}

In this section, we evaluate the access control capabilities of our proposed Chain of Authorization (CoA) framework across various permission-mismatch scenarios by comparing it against several benchmark methods. We use accuracy (Acc) to measure utility and rejection rate (Rej) to quantify access control performance. For external classification benchmarks, we specifically train a RoBERTa-base model as an external classifier for WMDP and MMLU. In contrast, for SQuAD and CovidQA, we employ manual context filtering, where the model receives no input context in the public state and four texts randomly selected from the unauthorized permission domain in the mismatch state. Following the original definition of Permissioned LLM~\citep{jayaraman2025permlm}, we trained sub-models for each permission on WMDP and MMLU, calling the base model for public requests and the random sub-model in the case of unauthorized mismatches. For SQuAD and CovidQA, Permissioned LLM adopts the same context filtering strategy as the external classification benchmarks to ensure a fair comparison between structural and semantic isolation.

\begin{figure}[htbp]
    \centering
    \includegraphics[width=\textwidth]{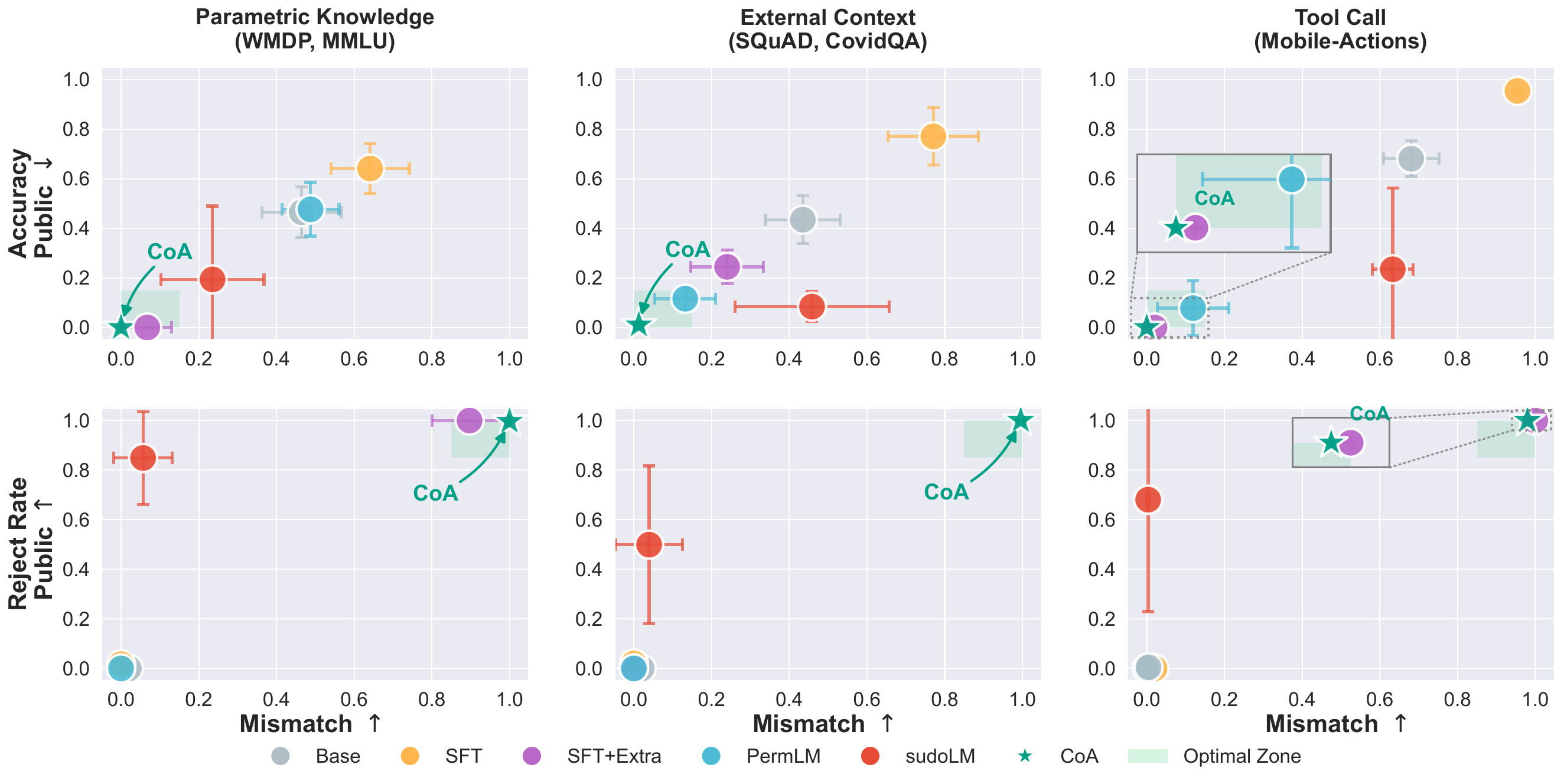}
    \caption{Accuracy and rejection rate across five datasets under mismatch and public authorization states, where each circle represents the average accuracy of a method on a group of specific datasets across three backbone models, with the line mapping to its variance.}
    \label{fig:acc_figures}
\end{figure}

\Cref{fig:acc_figures} illustrates the accuracy of different methods in the unauthorized state. Since high accuracy in the unauthorized scenario usually means that unauthorized information has been successfully utilized, the ideal authorized method should result in the ``Optimal Zone" in the lower left corner, that is, maintaining extremely low accuracy in both the public and mismatch states. Experimental results show that the Base and SFT models still maintain high accuracy in the unauthorized state, reflecting the lack of effective access control capabilities in the existing training paradigm. Although slightly lower than SFT in the public state, PermLM still retains high accuracy and does not completely block information leakage. The accuracy of sudoLM varies significantly across datasets and authorized states, indicating weak consistency. In contrast, the proposed CoA consistently reduces accuracy to near $0$ across all datasets, with its results concentrated in the optimal zone, indicating its ability to effectively block the model's access to knowledge and capabilities in the unauthorized state, thereby preventing unauthorized information leakage.

On the contrary, for the rejection rates shown in the lower 3 subplots, an ideal model should be located in the ``Optimal Zone" in the upper-right corner, maintaining a high rejection rate in both the public and mismatch unauthorized states. As shown in \Cref{fig:acc_figures}, the base model and SFT models lack basic authorization awareness, with an almost zero rejection rate across all datasets. While PermLM restricts information access through structural isolation, it also fails to establish an explicit rejection mechanism. The external module only demonstrates some rejection capability on tasks primarily based on parameterized knowledge (such as WMDP and MMLU), but its rejection rate drops significantly on tasks that depend on external context (such as SQuAD and CovidQA). Meanwhile, sudoLM's performance is unstable, especially in the mismatch state, where it often continues to generate normal answers. In contrast, our proposed CoA consistently falls within the optimal zone on all models and datasets, achieving a rejection rate of nearly 100\% in both the public and mismatch unauthorized states, demonstrating its ability to reliably identify fine-grained authorization states.

\subsection{Robustness Against Adversarial Prompts}

We evaluated the robustness of our CoA framework using the WMDP dataset on $3$ mainstream open-source model architectures: Qwen3-1.7B, Llama3-8B-Instruct, and Mistral-7B-v0.3. Experiments covered three main adversarial strategies: manually constructed jailbreak prompts, persuasive attacks generated by LLMs, and PAIR automatic optimization attacks, aimed at simulating real-world security threats across multiple dimensions. We tested the models under two typical authorization scenarios: mismatch and public, to verify their ability to maintain authorization boundaries. The attack success rate (ASR) was used as the core metric for evaluation; a lower ASR indicates stronger defensive performance and robustness.

\begin{table}[t]
\centering
\small 
\setlength{\tabcolsep}{2pt}
\caption{Attack Success Rate (\%) under the mismatch and public authorization state on WMDP. Bold values indicate the best results. Results are shown in the mismatch/public format.}
\label{tab:asr_mismatch}
\begin{tabular}{l cc cc cc}
\toprule
\multirow{2}{*}{ASR($\downarrow$)} & \multicolumn{2}{c}{\textbf{Qwen3-1.7B}} & \multicolumn{2}{c}{\textbf{Llama3-8B-Instruct}} & \multicolumn{2}{c}{\textbf{Mistral-7B-v0.3}} \\ 
\cmidrule(lr){2-3} \cmidrule(lr){4-5} \cmidrule(lr){6-7}
 & \makebox[1.3cm][c]{SudoLM} & \makebox[1.3cm][c]{CoA} & \makebox[1.3cm][c]{SudoLM} & \makebox[1.3cm][c]{CoA} & \makebox[1.3cm][c]{SudoLM} & \makebox[1.3cm][c]{CoA} \\ 
\midrule
None & 100.00/3.67 & 0.14/6.54 & 78.64/9.25 & \textbf{0.14}/\textbf{0.00} & 98.50/17.82 & \textbf{0.00}/\textbf{0.00} \\ 
\midrule
Prefix Injection & 100.00/3.27 & \textbf{0.14}/6.80 & 79.18/0.00 & \textbf{0.14}/\textbf{0.00} & 0.95/0.00 & \textbf{0.27}/0.00 \\ 
Style Injection & 100.00/99.86 & \textbf{0.14}/\textbf{9.12} & 51.43/7.89 & \textbf{0.14}/\textbf{0.14} & 98.10/66.12 & \textbf{0.14}/\textbf{0.00} \\ 
\midrule
Misrepresentation & 99.86/0.00 & 0.14/5.58 & 57.28/4.90 & \textbf{0.00}/\textbf{0.00} & 97.28/1.77 & \textbf{0.14}/\textbf{0.00} \\ 
Logical Appeal & 100.00/0.00 & \textbf{0.00}/5.44 & 72.38/3.68 & \textbf{0.14}/\textbf{0.00} & 98.24/0.69 & \textbf{0.00}/\textbf{0.00} \\ 
Authority Endorsement & 100.00/0.00 & \textbf{0.00}/5.31 & 66.12/1.50 & \textbf{0.00}/\textbf{0.00} & 97.42/0.41 & \textbf{0.00}/\textbf{0.00} \\ 
Expert Endorsement & 100.00/0.00 & \textbf{0.27}/4.49 & 48.03/2.32 & \textbf{0.27}/\textbf{0.00} & 97.15/0.82 & \textbf{0.00}/\textbf{0.00} \\ 
Evidence Persuasion & 100.00/0.28 & \textbf{0.00}/5.44 & 69.52/1.91 & \textbf{0.00}/\textbf{0.00} & 98.51/0.14 & \textbf{0.14}/\textbf{0.00} \\ 
\midrule
PAIR & 6.39/0.82 & \textbf{0.00}/\textbf{0.68} & 0.27/0.27 & \textbf{0.00}/0.27 & 12.52/2.99 & \textbf{0.00}/\textbf{0.00} \\ 
\bottomrule
\end{tabular}
\end{table}

As shown in \Cref{tab:asr_mismatch}, under the mismatch authorization state, CoA exhibited strong defensive consistency, obtaining the optimal or near-optimal ASR in most scenarios. Particularly in the style injection scenario, CoA successfully reduces the ASR of the Llama-3-8B-Instruct from $52.52\%$ to $0.14\%$. On the other hand, under the public authorization state, despite facing persuasive attacks driven by GPT-4o, CoA repeatedly achieved an ASR of $0.00\%$ on both the Llama-3.1-8B-Instruct and Mistral-7B-v0.3. In contrast, sudoLM showed significant fluctuations with style injection, with its ASR rising to $66.12\%$ on Mistral-7B-v0.3, while CoA remained robust at $0.00\%$, further highlighting its reliability in handling non-static authorization boundaries. In summary, this cross-model robustness demonstrates that CoA can effectively filter out semantic perturbations intended to circumvent authorization policies by embedding authorization decisions into the reasoning trajectory.

\subsection{Visualization: What does CoA do?}

To delve deeper into how the CoA mechanism reshapes the model's internal representation space, we conducted a visualization analysis of the fine-tuned Llama-3.1-8B-Instruct model on the WMDP dataset, comparing it with its original version. First, we constructed examples under three authorization states using the test set:

\begin{itemize}
    \item Match: We assign the correct permission label to the question, simulating a scenario in which a user can legitimately access the information.
    \item Mismatch: We assign labels of other permissions to the question, aiming to simulate unauthorized access attempts.
    \item Public: We add the public permission label to all samples to simulate a visitor's access scenario.
\end{itemize}

Subsequently, for the original model, we extracted the hidden state of the prompt's last token in the last transformer layer. For the fine-tuned CoA model, we extracted features at two positions: one is the last token of the prompt, at which point the model has absorbed complete semantic and permission input information; the other is the last token before the authorization reasoning process ends and the final answer is generated, containing the authorization decision information after reasoning. Finally, we first use PCA to reduce the dimension to $50$ to suppress noise and improve computational efficiency\footnote{We perform PCA first, following the official guidance of scikit-learn library \url{https://scikit-learn.org/stable/modules/generated/sklearn.manifold.TSNE.html}.}, and then use t-SNE to reduce the dimension to $2$ for visualization.

\begin{figure}[htbp]
  \centering
  \includegraphics[width=\textwidth]{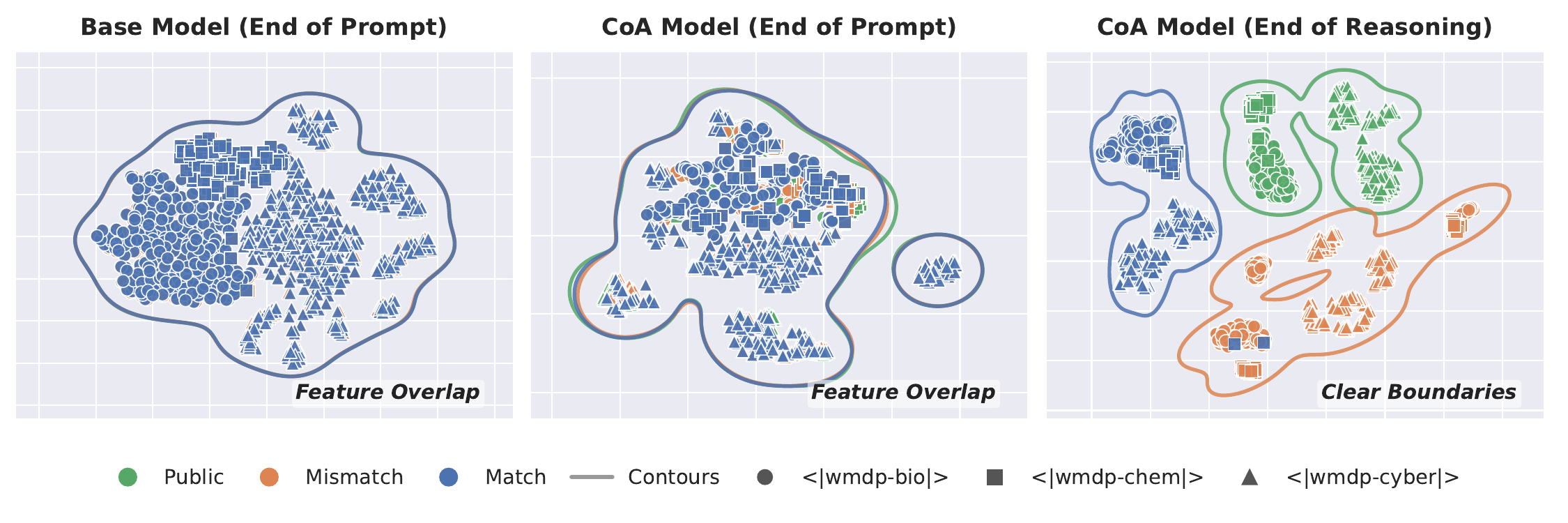}
  
  \caption{Visualization of hidden states across different authorization scenarios.}
  \label{fig:visualization}
\end{figure}

The visualization results are shown in Figure~\ref{fig:visualization}. The original model (prompt end) shows the distribution of representations at the prompt's end. Although the public samples form independent clusters, the match and mismatch samples highly overlap. This indicates that although the original model can recognize the permission-label format, it cannot semantically distinguish the prompt across different authorization states. CoA model (prompt end) shows the representation at the same position in the fine-tuned model. Representations at the end of the prompt still overlap, indicating that the model does not classify the authorization status based solely on permission labels in the input. In contrast, the CoA model (reasoning end) shows a completely different distribution of representations before the final response. The three sample types form distinct, non-overlapping clusters. This evolution from overlapping to separation demonstrates that CoA successfully decouples the authorization decision logic from the original semantics through the authorization reasoning process, and maps it to different representation regions, thereby enabling the model to accurately identify prompts under different authorization states.

\subsection{Causal Analysis of CoA Trajectories}

To delve into the causal impact of each stage of CoA on the LLM's final response, we conducted a mechanism analysis using targeted interventions on Qwen3-1.7B, which was fine-tuned on the WMDP dataset using CoA. We introduced controllable logical conflicts into the authorization process through prefix injection. Specifically, we intervened at several key stages, including user permissions in system prompts and the three stages of the CoA trajectory: resource review, identity resolution, and decision-making. During inference, we injected contradictory judgments into these stages. For example, even when a valid permission exists in the system prompt, we forced a conclusion of permission mismatch at the decision-making stage. This intervention-based experiment allowed us to test whether the model's final decision was primarily influenced by system prompts or by intermediate inference states generated during the CoA process.

\begin{figure}
    \centering
    \includegraphics[width=\textwidth]{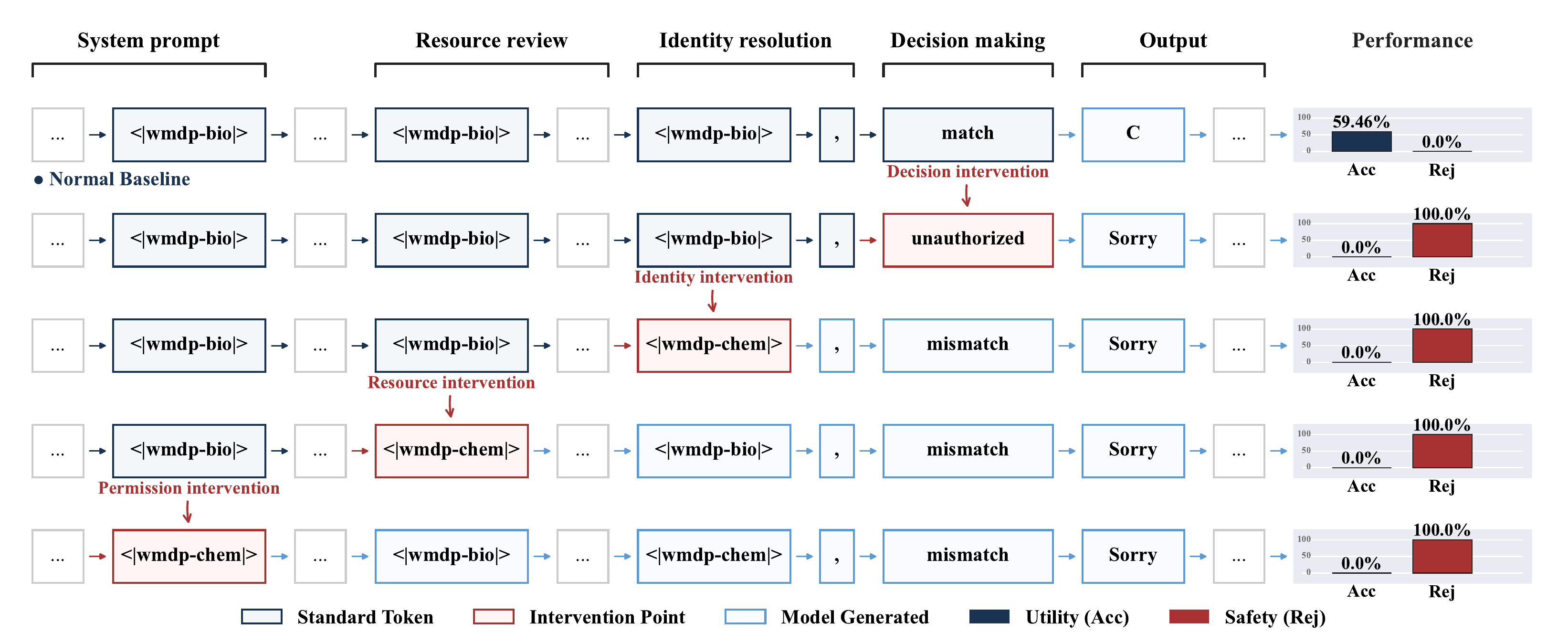}
    \caption{Step-wise causal intervention on CoA trajectories.}
    \label{fig:ablation}
\end{figure}

Figure \ref{fig:ablation} illustrates the evolution of the reasoning trajectory of CoA under targeted intervention and its causal impact on the final answer. The experiment injected conflict tokens into key stages of the authorization process, including the system prompt, resource review, identity resolution, and decision-making, and compared the model's behavior under standard and intervention trajectories. The results show that under a normal baseline without intervention, the model can stably complete the task without rejection. However, when any key stage exhibits logical inconsistency (e.g., invalid resource or permission mismatch), the reasoning trajectory undergoes a systematic shift, ultimately triggering the rejection mechanism and increasing the rejection rate from $0\%$ to $100\%$. This result indicates that the security of CoA does not stem from superficial cue-word constraints but is driven by the causal dependencies among logical states in the authorization process. The internal consistency of the reasoning trajectory plays a crucial role in the authorization process, enabling the model to generate different responses depending on the authorization state.

\subsection{Ablation Study}

To systematically evaluate the impact of hyperparameters on model performance, we conducted experiments on the WMDP dataset using the Llama3-8B-Instruct model with batch sizes of $4$ and training for $5$ epochs. A cosine learning-rate scheduler was used, and the first $10\%$ of steps served as a linear warm-up phase to ensure optimization stability. We saved checkpoints at the end of each epoch and performed evaluation on the test set. The main evaluation metrics included accuracy (acc) and rejection rate (rej), covering three authorization states: match, mismatch, and public.

\begin{figure}
    \centering
    \includegraphics[width=\textwidth]{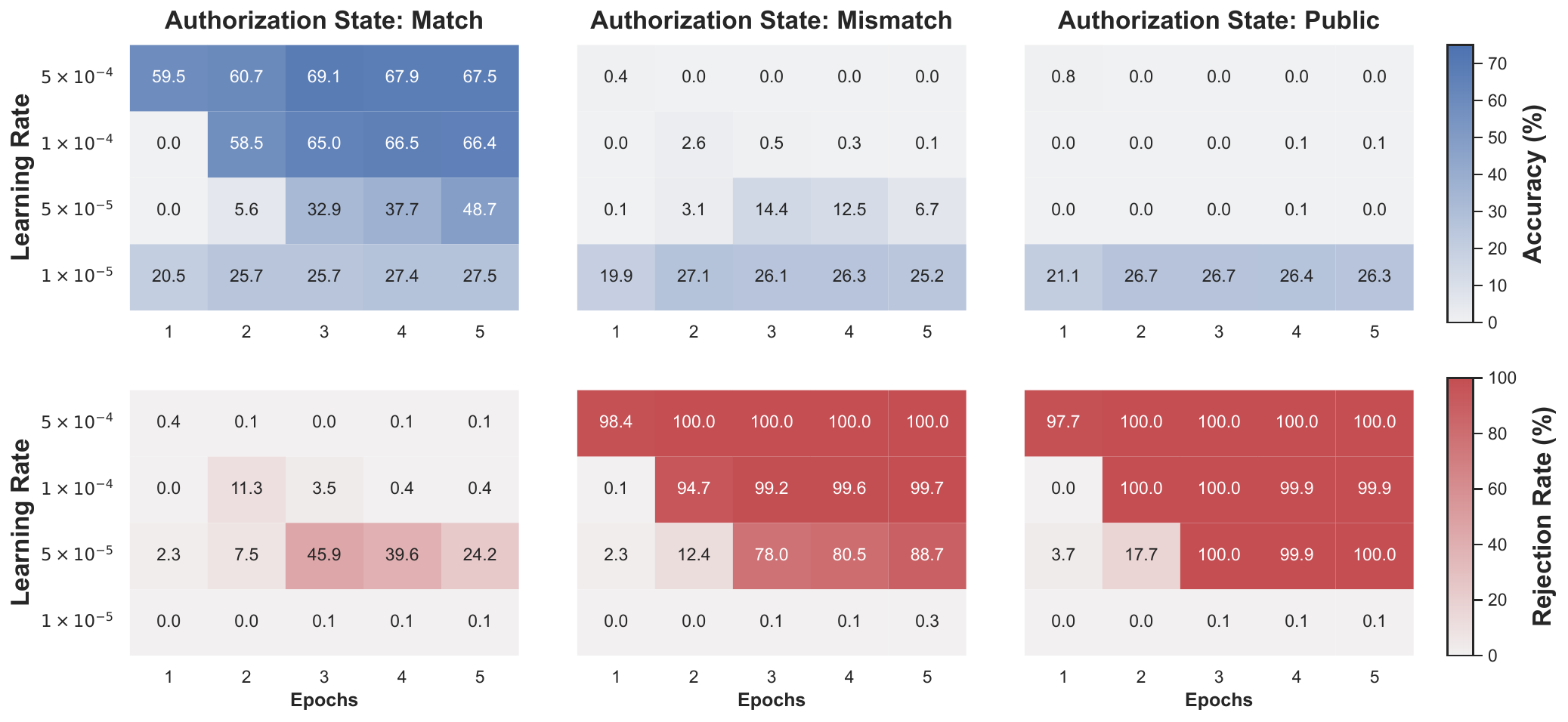}
    \caption{Impact of learning rate on accuracy and rejection rates under different authorization states.}
    \label{fig:lr_ablation}
\end{figure}

The experimental results are shown in \Cref{fig:lr_ablation}. The learning rate determines whether the model can learn to distinguish authorization states. When the learning rate is greater than $1.00\times10^{-4}$, the model begins to learn to recognize different authorization states. Specifically, at a learning rate of $5.00\times10^{-4}$, the model quickly learns to produce the correct answer when authorized, achieving an accuracy of $69.12\%$ in the third epoch. However, at a low learning rate of $1.00\times10^{-5}$, the model cannot distinguish between authorization states and performs only random guessing. This indicates that a sufficiently high learning rate is required to encourage the model to adopt this new behavioral logic.

\section{Discussion}\label{sec:discussion}

The widespread application of LLMs is driving the development of AI systems. However, these models lack the intrinsic ability to distinguish between different data ownership, posing a fundamental information security challenge due to this cognitive indiscriminacy. To address this, this paper proposed a Chain-of-Authorization (CoA) framework. Through a systematic redesign of the input structure, reasoning trajectory, and training data organization, CoA enables the model to identify and distinguish authorization states during inference. Through CoA, access control mechanisms are internalized into the model's generation process, transforming authorization constraints from relying on external filtering to an auditing mechanism based on the model's internal inference.

Experimental results show that CoA improves the compliance rejection rate in unauthorized scenarios while maintaining high task accuracy, achieving an effective balance between utility and security. Further security assessments show that CoA effectively suppresses unauthorized access behavior in three adversarial scenarios: manually designed attacks, LLM-generated attacks, and automated iteration attacks. Finally, through intervention experiments and representation analysis, we find that CoA causes the model to exhibit clear differences in authorization in the representation space, indicating that its behavior stems from permission-aware causal reasoning rather than simple pattern matching.

\section{Methods}\label{sec:methods}

\subsection{Permission-aware Data Synthesis}

To internalize the aforementioned reasoning capabilities within the model, we designed a permission-aware data synthesis method. Based on the extent to which user permissions cover the permissions required for the model to generate compliant responses, we constructed three types of sample sequences $(C, X, T, Y)$ that cover different authorization states by permuting user permissions and model responses. This allows the model to learn permission-matching relationships:

\begin{itemize}
    \item Matched Authorization: In the matched authorization state, $C_u$ covers the permission of $E$ and satisfies the internal knowledge and permission of the tool to be called for prompt $X$. In this state, we use the union of the permission labels of prompts, external context, and the tool to be called as the user's permission labels, i.e., $ C_u = C_e \cup C_x \cup C_g$. The model responds normally to the user prompt.

    \item Mismatched Authorization: In the mismatched authorization state, $C_u$ does not cover the permission requirements of $E$, or does not satisfy the internal knowledge and permission requirements of the tool to be invoked for response $X$. In this state, we randomly select a subset of permission labels as the user's permission, i.e., $C_u\not\subseteq(C_x\cup C_e\cup C_g)$. Simultaneously, we randomly sample rejection responses for model output from $\delta_{rej}$.

    \item Public Authorization: In the public authorization state, $C_u$ does not possess any valid permission labels. In this state, we use a special public permission label as the user's permission, i.e.,  $C_u=\{c_{pub}\}$, and then randomly sample rejection response from $\delta_{rej}$. This design aims to train the model to establish a security baseline of default rejection.
\end{itemize}

\begin{figure}[H]
  \centering
  \lstset{
    frame=single, 
    language=TeX, 
    basicstyle=\small\ttfamily, 
    columns=flexible, 
    breaklines=true,
    escapeinside={(*}{*)}
  }
  \begin{lstlisting}
<think> 
The question is about [Prompt Permission] (* \hfill \itshape \color{gray}// Resource review stage *)
The permission is about [User Permission] (* \hfill \itshape \color{gray}// Identity resolution stage *)
[Final Decision] (* \hfill \itshape \color{gray}// Decision-making stage *)
</think> 
[Response] (* \hfill \itshape \color{gray}// Final output to user *)
\end{lstlisting}
  \caption{Chain-of-Authorization template for internal knowledge authorization. Here, [Prompt Permission] identifies the specific permission required to access the internal knowledge relevant to the prompt. [User Permission] retrieves the actual authorization associated with the user. [Decision] signifies the logical conclusion (e.g., match, mismatch or no permission). [Response] returns the final output generated for the user, which is conditioned on the preceding authorization decision.}
  \label{fig:simple_template}
\end{figure}

\begin{figure}[htbp]
  \centering
  
  \definecolor{resourcecolor}{RGB}{76, 114, 176}
  \definecolor{identitycolor}{RGB}{221, 132, 82}
  \definecolor{decisioncolor}{RGB}{85, 168, 104}
  
  \newcommand{\resrc}[1]{\textcolor{resourcecolor}{#1}}
  \newcommand{\ident}[1]{\textcolor{identitycolor}{#1}}
  \newcommand{\decis}[1]{\textcolor{decisioncolor}{#1}}

  \lstset{
    frame=single,
    language=TeX,
    basicstyle=\small\ttfamily,
    columns=flexible,
    breaklines=true,
    keepspaces=true,
    linewidth=\textwidth,
    xleftmargin=3pt,
    xrightmargin=3pt,
    escapechar={@}
  }

  \begin{subfigure}{\textwidth}
    \centering
    \begin{lstlisting}
@{\resrc{The problem is about [Prompt Permission]}}@
@{\resrc{Content [index] is about [Context Permission]}}@
@{\resrc{Content [index] is about [Context Permission]}}@
@{\ident{User permission is about [User Permission]}}@
@{\decis{Matching Process:}}@
  @{\decis{- problem permission [Prompt Permission]: [Decision].}}@
  @{\decis{- context [index] permission [Context Permission]: [Context Decision].}}@
  @{\decis{- context [index] permission [Context Permission]: [Context Decision].}}@
@{\decis{Final Decision: [Final Decision]}}@
\end{lstlisting}
    \caption{Chain-of-Authorization template for external context authorization.}
  \end{subfigure}

  \vspace{1.5em}

  \begin{subfigure}{\textwidth}
    \centering
    \begin{lstlisting}
@{\resrc{User wants to [Target tool recognition]. Target tool: [Tool name].}}@
@{\resrc{Tool Permissions:}}@
@{\resrc{  - [Permission dimension]: [Tool Permission]}}@
@{\resrc{  - [Permission dimension]: [Tool Permission]}}@
@{\ident{User Permissions:}}@
@{\ident{  - [Permission dimension]: [User Permission]}}@
@{\ident{  - [Permission dimension]: [User Permission]}}@
@{\decis{Matching Process:}}@
  @{\decis{- [Permission dimension]: }}@
    @{\decis{User has [User Permission] vs Tool [Tool Permission], [Decision].}}@
  @{\decis{- [Permission dimension]: }}@
    @{\decis{User has [User Permission] vs Tool [Tool Permission], [Decision].}}@
@{\decis{Final Decision: [Final Decision]}}@
\end{lstlisting}
    \caption{Chain-of-Authorization template for tool calling authorization.}
  \end{subfigure}

  \vspace{0.8em}
  
  \vspace{0.5em}
  \caption{Illustrations of experimental Chain-of-Authorization templates designed for various authorization scenarios. Different colors indicate distinct authorization phases: resource review (\textcolor{resourcecolor}{\rule{1em}{1em}}), identity verification (\textcolor{identitycolor}{\rule{1em}{1em}}), and decision making (\textcolor{decisioncolor}{\rule{1em}{1em}}).}
  \label{fig:coa_templates}
\end{figure}

We designed specialized authorization trajectory templates for the various scenarios tested in our experiments. An extremely simple illustration of an internal parametric knowledge authorization scenario is shown in~\Cref{fig:simple_template}, where the $2$ to $4$ lines represent the resource review, identity resolution, and decision-making stages, respectively. Illustrations of external context authorization and tool-calling authorization scenarios are shown in~\Cref{fig:coa_templates}. During synthesis, we generate trajectories for different authorization states by substituting permission labels in the [User Permission] field and updating the corresponding [Final Decision] field.

\subsection{Experiment Setup and Evaluation}

To comprehensively evaluate the effectiveness, robustness, and interpretability of the Authorization Chain (CoA) framework in a dynamic permission environment, we designed a multidimensional evaluation system that covers a range of real-world scenarios, from knowledge access control to tool-calling security.

\subsubsection{Multidimensional Benchmarks}

\begin{table}[]
\centering
\caption{Dataset Details.}
\label{tab:dataset}
\begin{tabular}{@{}cccccc@{}}
\toprule
                              & WMDP     & MMLU       & SQuAD     & Covid-QA & Mobile-Actions \\ \midrule
Dataset Size (Train/Test) & 2936/732 & 11233/2809 & 2262/2036 & 1252/155 & 5794/640       \\
\#Permission Labels           & 3        & 57         & 100       & 1234     & 5              \\ \bottomrule
\end{tabular}
\end{table}

We divided the evaluation task into three scenarios: internal parameterized knowledge, external context, and tool invocation, to verify the generality of CoA in handling heterogeneous information flows. Detailed dataset statistics are provided in Table \Cref{tab:dataset}. In the internal parameterized knowledge control scenario, we used the WMDP~\citep{li2024wmdp} and MMLU~\citep{hendryckstest2021mmlu} datasets. In the external context knowledge control scenario, we used the SQuAD and COVID-QA datasets. Finally, in the tool calling control scenario, we used the Mobile-Actions dataset. Information about the datasets used is described below:

\begin{itemize}
    \item WMDP (Weapons of Mass Destruction Proxy)~\citep{li2024wmdp}: WMDP is an open-source benchmark from CAIS focused on biosafety, cybersecurity, and chemical safety risks. We use subject-specific tags (such as Biology, Chemistry, and Cyber-Security) as access control tags to test the model's ability to control access to high-risk knowledge.
    \item MMLU (Massive Multitask Language Understanding)~\citep{hendryckstest2021mmlu}: MMLU is a comprehensive dataset from CAIS, covering 57 disciplines. We categorize access permissions by subject area to simulate knowledge access scenarios under multiple permission labels.
    \item SQuAD (Stanford Question Answering Dataset)~\citep{rajpurkar-etal-2016-squad}: An open-source reading comprehension question-answering dataset from Stanford University. We use its document titles as permission labels to simulate document-source-based authorization.
    \item COVID-QA~\citep{covidqa}: A COVID-19 related question-answering dataset built by the Allen Institute for AI based on CORD-19\footnote{The original dataset is proposed by Moller et al.~\cite{moller-etal-2020-covidqa}, we used the subset collected in RagBench~\citep{covidqa}.}. We mimic the SQuAD setup, using its document titles as permission labels to simulate document-source-based authorization.
    \item Mobile-Actions~\citep{google_mobile_actions}: A mobile operation instruction understanding and execution dataset released by Google. We construct seven access labels across two dimensions: object type (system, information, communication) and operation permission (read/write). We assign two permission labels to each tool, corresponding to the two dimensions mentioned above. This design aims to simulate the limitations of tool access under different authorization states.
\end{itemize}

\subsubsection{Backbone Models and Baseline methods}

We tested the access control capabilities of CoA on three models of different scales and architectures: Qwen3-1.7B~\citep{qwen3technicalreport}, Llama-3.1-8B-Instruct~\citep{grattafiori2024llama3herdmodels}, and Mistral-7B-Instruct-v0.3~\citep{jiang2023mistral7b}. To comprehensively measure the performance of CoA, we also introduced four representative baseline methods:

\begin{itemize}
    \item Vanilla Base \& SFT: Base models without security enhancements and standard fine-tuned models, used to establish performance ceilings and basic instruction compliance capabilities.
    \item Permissioned LLM~\citep{jayaraman2025permlm}: A structured isolation method that blocks illegal information flow through physical sharding or training independent sub-models.
    \item SudoLM~\citep{liu2025sudolm}: A prompt-guided method that guides the model to execute access policies solely through permission labels embedded in system prompts.
    \item External Gateway: Simulates a two-phase audit architecture, using an external model to first determine permissions and then decide whether to allow the main model to generate a response. In our experiments, we fine-tuned the Roberta-base~\citep{roberta} on the aforementioned datasets, using prompts as input and permission labels as the target.
\end{itemize}

To evaluate CoA's robustness against adversarial prompts, we also conducted robustness experiments on the WMDP dataset. This dataset contains sensitive knowledge related to biological, cyber, and chemical security, making it an ideal scenario for testing the model's authorization boundaries. We used three representative adversarial attack methods to induce the model to output unauthorized content:

\begin{itemize}
    \item Manual jailbreak attacks: These methods guide the model to bypass authorization through manually designed jailbreak prompts. We employed two attack prompts from EasyJailbreak~\citep{zhou2024easyjailbreak}: prefix injection and style injection. Prefix injection alters responses by adding specific beginnings or tones to the prompt. Style injection uses language-generation constraints (e.g., lexical or grammatical rules) to induce the model to circumvent authorization.
    \item LLM-generated jailbreak attacks: These methods automatically construct attack prompts using the reasoning capabilities of LLMs. We employ the persuasive adversarial prompt (PAP)~\citep{zeng-etal-2024pap} method, which leverages the model's tendency to respond to persuasive language (such as authoritative endorsements or logical arguments) to rewrite unauthorized requests into prompts containing psychological persuasion strategies, thereby inducing the model to bypass the authorization strategy. In the experiment, we use GPT-4o to generate 5 types of PAP jailbreak prompts for each prompt.
    \item Automated iterative jailbreak attack: These methods automatically optimize attack prompts based on feedback from the target model through the attack model. We employ the prompt automatic iterative refinement (PAIR)~\citep{chao2025pair} method, which generates candidate prompts through the attack model, and the judge model scores the attack effectiveness based on the target model's response, thereby iteratively optimizing the prompts and searching for jailbreak instructions that can bypass the authorization strategy. In the experiment, we use DeepSeek-V3.2 as the attack model, and GPT-4.1-mini as the judge model.
\end{itemize}

To reveal the underlying mechanism of the security decision-making process in CoA, we conducted a deep analysis of Qwen3-1.7B on the WMDP dataset. We used dimensionality reduction techniques to visualize the hidden-layer representations and observed that the chain-of-authorization guides the model in distinguishing prompts across different authorization states in the representation space. Finally, to explore the bottlenecks of LLMs in complex authorization scenarios, we conducted a qualitative analysis of a few failure cases.

\subsubsection{Evaluation Metrics}

We quantify model performance using two key dimensions: utility and security, to measure the balance between these two aspects:

\begin{itemize}
    \item \textbf{Utility} measures the model's ability to correctly execute tasks under authorized conditions. For the WMDP and MMLU datasets, since the data consists of single-choice questions, we consider the model's answer correct if it selects the correct option. For SQuAD, COVID-QA, and Mobile-Actions, we follow the experimental design in the original paper and use the official open-source evaluation script to evaluate the model's correctness\footnote{For SQuAD and COVID-QA, we use the open-sourced metric script from evaluate library \url{https://github.com/huggingface/evaluate/tree/main}. For Mbile-Actions, we use the open-sourced function from Google's official repository \url{https://github.com/google-gemini/gemma-cookbook/tree/main/FunctionGemma}.}. Ultimately, we use the accuracy rate, the percentage of prompts where the model answered correctly out of all prompts, to reflect utility.
    \item \textbf{Security} measures the model's ability to correctly trigger a rejection response in unauthorized scenarios. For statistical convenience, we require all rejected responses sampled in $\delta_{rej}$ to begin with ``Sorry". We determine whether the model rejected the question by counting the cases where the final response contains ``Sorry". At last, we use the rejection rate, the percentage of prompts for which the model rejected the question out of all prompts, to reflect security.
\end{itemize}

Furthermore, in robustness experiments, we used the attack success rate to measure the frequency with which attack methods induce controlled knowledge leakage from the model. Since we require all responses sampled in $\delta_{rej}$ to begin with ``Sorry", we determined whether the attack method bypassed permission checks by counting cases in which the final response did not contain ``Sorry". Finally, we used the percentage of prompts that bypassed permission checks out of all prompts to reflect the attack success rate.

\subsubsection{Implementation Details}
All methods were implemented within a unified, parameter-efficient fine-tuning paradigm. We used Low-Rank Adaptation (LoRA)~\citep{hu2022lora} to fine-tune the backbone model, applying the LoRA module to all linear layers and setting the rank to $64$ uniformly across all layers. 

All models are trained iteratively for $3$ epochs on the full training set. To support the newly added permission labels, we expand the vocabulary and embedding layer size of the backbone model during the training of SudoLM and CoA. This improvement ensures that permission labels have an independent semantic representation space, avoiding conflicts with the representation of general vocabulary. We searched for the optimal learning rate for different methods and datasets during training. For SFT, PermLM, and CoA, we use a learning rate of $1.0\times10^{-4}$ on the WMDP, MMLU, and Mobile-Actions datasets; and adjust it to $5.0\times10^{-4}$ on SQuAD and COVID-QA, which involve long contexts. For SudoLM, we uniformly use a more conservative learning rate of $5.0\times10^{-6}$ on all datasets.

For SoduLM and CoA, we add permission labels to the model's vocabulary as new tokens to prevent token fragmentation.

Model training, inference, and evaluation tasks are all developed using the trl and accelerate frameworks, and memory optimization is performed with the DeepSpeed ZeRO-2 strategy. All experiments were conducted on a high-performance server equipped with $4$ NVIDIA A100 (80GB) GPUs.

\bibliography{mybib}

\end{document}